\documentclass[runningheads]{llncs}

\usepackage{graphicx}
\usepackage{hyperref}
\usepackage{xcolor}
\usepackage{textcomp}
\usepackage{bbding}
\usepackage{amsfonts}
\usepackage{amsmath}
\usepackage{amssymb}
\usepackage{mathtools}
\usepackage{mwe}

\def\bag{{\cal B}}
\def\prenet#1#2{\,\ensuremath{\stackrel{#1}{\bullet}\!{#2}}}
\def\postnet#1#2{\ensuremath{{#2}\!\kern-.05ex\stackrel{#1}{\bullet}}\,}
\def\pre#1{\ensuremath{\bullet{#1}}}
\def\post#1{\ensuremath{{#1}\kern-.05ex\bullet}}
\def\mi#1{\mathit{#1}}
\def\nomove{\gg}

\newcommand{\Nat}{\ensuremath{\mathrm{I\kern-1.5pt N}}}

\hypersetup{
	colorlinks,
	linkcolor={red!50!black},
	citecolor={blue!60!black},
	urlcolor={blue!80!black}
}
\begin{document}

\title{Mining Uncertain Event Data\\ in Process Mining\thanks{In \emph{International Conference on Process Mining} (ICPM 2019). DOI: 10.1109/ICPM.2019.00023. \textcopyright IEEE. We thank the Alexander von Humboldt (AvH) Stiftung for supporting our research interactions. Please do not print this document unless strictly necessary.}}

\author{Marco Pegoraro~\Envelope\orcidID{0000-0002-8997-7517} \and
	Wil M.P. van der Aalst\orcidID{0000-0002-0955-6940}}

\authorrunning{Pegoraro et al.}

\institute{Process and Data Science Group (PADS) \\ Department of Computer Science, RWTH Aachen University, Aachen, Germany
	\email{\{pegoraro,wvdaalst\}@pads.rwth-aachen.de}\\
	\url{http://www.pads.rwth-aachen.de/}}

\maketitle

\begin{abstract}
Nowadays, more and more process data are automatically recorded by information systems, and made available in the form of \emph{event logs}. Process mining techniques enable process-centric analysis of data, including automatically discovering process models and checking if event data conform to a certain model. In this paper we analyze the previously unexplored setting of uncertain event logs: logs where quantified uncertainty is recorded together with the corresponding data. We define a taxonomy of uncertain event logs and models, and we examine the challenges that uncertainty poses on process discovery and conformance checking. Finally, we show how upper and lower bounds for conformance can be obtained aligning an uncertain trace onto a regular process model.

\end{abstract}

\section{Introduction}
Over the last decades, the concept of \emph{process} has become more and more central in formally describing the activities of businesses, companies and other similar entities, structured in specific steps and phases. A process is thus defined as a well-structured set of activities, possibly performed by multiple actors (\emph{resources}), which contribute to the completion of a specific task or to the achievement of a specific goal.

The processes that govern the innards of business companies are increasingly supported by software tools. Performing specific activities is both aided and recorded by \emph{process-aware information systems} (PAISs), which support the de\-finition and management of processes. The information regarding the execution of processes, which includes time, case and activity information, can then be extracted from PAISs in the form of an \emph{event log}, a database or file containing the digital trace of the operations carried out in the context of the execution of a process and recorded as \emph{events}. The discipline of \emph{process mining} concerns the automatic analysis of event logs, with the goal of extracting knowledge regarding e.g. the structure of the process, the conformity of events to a specific normative process model, the performances in executing the process, the relationships between groups of actors in the process.

In this paper we will consider the analysis of a specific class of event logs: the logs that contain \emph{uncertain event data}. Uncertain events are recordings of executions of specific activities in a process which are enclosed with an indication of uncertainty in the event attributes. Specifically, we consider the case where the attributes of an event are not recorded as a precise value but as a range or a set of alternatives.

The recording of uncertain event data is a common occurrence in process management. The \emph{Process Mining Manifesto}~\cite{van2011process} describes a fundamental pro\-perty of event data as \emph{trustworthiness}, the assumption that the recorded data can be considered correct and accurate. In a general sense, uncertainty as defined here is an explicit absence of trustworthiness, with an indication of uncertainty recorded together with the event data. In the taxonomy of event data proposed in the Manifesto the logs at the two lower levels of quality frequently lack trustworthiness, and thus can be uncertain. This encompasses a wide range of processes, such as event logs of document and product management systems, error logs of embedded systems, worksheets of service engineers, and any process recorded totally or partially on paper.
There are many possible causes behind the recording of uncertain event data, such as:
\begin{itemize}
	\item \emph{Incorrectness}. In some instances, the uncertainty is simply given by errors occurred while recording the data itself. Faults of the information system, or human mistakes in a data entry phase can all lead to missing or altered event data that can be subsequently modeled as uncertain event data.
	\item \emph{Coarseness}. Some information systems have limitations in their way of recording data - often tied to factors like the precision of the data format - such that the event data can be considered uncertain. A typical example is an information system that only records the date, but not the time, of the occurrence of an event: if two events are recorded in the same day, the order of occurrence is lost. This is an especially common circumstance in the processes that are, partially or completely, recorded on paper and then digitalized. Another factor that can lead to uncertainty in the time of recording is the information system being overloaded and, thus, delaying memorization of data. This type of uncertainty can also be generated by the limited sensibility of a sensor.
	\item \emph{Ambiguity}. In some cases, the data recorded is not an identifier of a certain event attribute; in these instances, the data needs to be interpreted, either automatically or manually, in order to obtain a value for the event attribute. Uncertainty can arise if the meaning of the data is ambiguous and cannot be interpreted with precision. Example are data in the form of images, text, or video.
\end{itemize}
Aside from the causes, we can individuate other types of uncertain event logs based on the frequency of uncertain data. Uncertainty can be \emph{infrequent}, when a specific attribute is only seldomly recorded together with explicit uncertainty; the uncertainty is rare enough that uncertain events can be considered outliers. Conversely, \emph{frequent} uncertain behavior of the attribute is systematic, pervasive in a high number of traces, and thus not to be considered an outlier. The uncertainty can be considered part of the process itself. These concepts are not meant to be formal, and are laid out to distinguish between logs that are still processable regardless of the uncertainty, and logs where the uncertainty is too invasive to analyze them with existing process mining techniques.

In this paper we propose a taxonomy of the different types of explicit uncertainty in process mining, together with a formal, mathematical formulation. As an example of practical application, we will consider the case of conformance checking~\cite{carmona2018conformance}, and we will apply it to uncertain data by assessing what are the upper and lower bounds on the conformance score for possible values of the attributes in an uncertain trace.

The rest of this paper is organized as follows. Section~\ref{sec:related} discusses previous and related work in the management of uncertain data. Section~\ref{sec:taxonomy} proposes a taxonomy of the different possible types of uncertain process data. Section~\ref{sec:definitions} contains the formal definitions needed to manage uncertainty. Section~\ref{sec:conformance} describes a practical application of process mining over uncertain event data, the case of conformance checking through alignments. Section~\ref{sec:experiments} shows experimental results on computing conformance checking scores for uncertain data. Finally, Section~\ref{sec:conclusion} concludes the paper and discusses about future work.

\section{Related Work}\label{sec:related}
As mentioned, the occurrence of data containing uncertainty - in a broad sense - is common both in more classic disciplines like statistics and Data Mining~\cite{han2011data} and in process mining~\cite{van2011process}; and logs that show an explicit uncertainty in the control flow perspective can be classified in the lower levels of the quality ranking proposed in the process mining manifesto.

Within process mining there exist various techniques to deal with a kind of uncertainty different from the one that we analyze here: missing or incorrect data. This can be considered as a form of non-explicit uncertainty: no measure or indication on the nature of the uncertainty is given in the event log. The work of Suriadi et al.~\cite{suriadi2017event} provides a taxonomy of this type of issues in event logs, laying out a series of data patterns that model errors in process data. In these cases, and if this behavior is infrequent enough to allow the event log to remain meaningful, the most common way for existing process mining techniques to deal with missing data is by filtering out the affected traces and performing discovery and conformance checking on the resulting filtered event log. While filtering out missing values is straightforward, various methodologies of event log filtering have been proposed in the past to solve the problem of incorrect event attributes: the filtering can take place thanks to a reference model, which can be given as process specification~\cite{wang2015cleaning}, or from information discovered from the frequent and well-formed traces of the same event log; for example extracting an automaton from the frequent traces~\cite{conforti2017filtering}, computing conditional probabilities of frequent sequences of activities~\cite{sani2017improving}, or discovering a probabilistic automaton~\cite{van2018filtering}. In the latter cases, the noise is identified as infrequent behavior.

Some previous work attempt to repair the incorrect values in an event log. Conforti et al.~\cite{conforti2018timestamp} propose an approach for the restoration of incorrect timestamps based on a log automaton, that repairs the total ordering of events in a trace based on correct frequent behavior. Fani Sani et al.~\cite{sani2018repairing} define outlier behavior as the unexpected occurrence of an event, the absence of an event that is supposed to happen, and the incorrect order of events in the trace; then, they propose a repairing method based on probabilistic analysis of the context of an outlier (events preceding or following the anomalous event). Again, both of these methods define anomalous/incorrect behavior on the basis of the frequency of occurrence.

The main driving reasons behind this work is to provide the means to treat uncertainty as a relevant part of a process; thus, we aim not to filter it out but model it. In conclusion, there are two novel aspects regarding uncertain data that we intend to address in this work. The first is the \emph{explicitness of uncertainty}: we work with the underlying assumption that the actual value of the uncertain attribute, while not directly provided, is described formally. This is the case when meta-information about the uncertainty in the attribute is available, either deduced from the features of the information system(s) that record the logs or included in the event log itself. Note that, as opposed to all previous work on the topic, the fact that uncertainty is explicit in the data means that the concept of uncertain behavior is completely separated from the concept of infrequent behavior. The second is the goal of \emph{modeling uncertainty}: we consider uncertainty part of the process. Instead of filtering or cleaning the log we introduce the uncertainty perspective in process mining by extending the currently available techniques to incorporate it.

\section{A Taxonomy of Uncertain Event Data}\label{sec:taxonomy}
The goal of this section of the paper is to propose a categorization of the different types of uncertainty that can appear in process mining. In process management, a central concept is the distinction between the data perspective (the event log) and the behavioral perspective (the process model). The first one is a static representation of process instances, the second summarizes the behavior of a process. Both can be extended with a concept of explicit uncertainty: this concept also implies an extension of the process mining techniques that have currently been implemented.

In this paper we will focus on uncertainty in event data, while the concept of uncertainty applied to models will be examined in a future work. Specifically, as an example application we will consider computing the conformance score of uncertain process data on classical models.

We can individuate two different notions of uncertainty:

\begin{itemize}
	\item \emph{Strong uncertainty}: the possible values for the attributes are known, but the probability that the attribute will assume a certain instantiation is unknown or unobservable.
	\item \emph{Weak uncertainty}: both the possible values of an attribute and their respective probabilities are known.
\end{itemize}

\begin{figure}
	\centering
	\includegraphics[width=1\columnwidth]{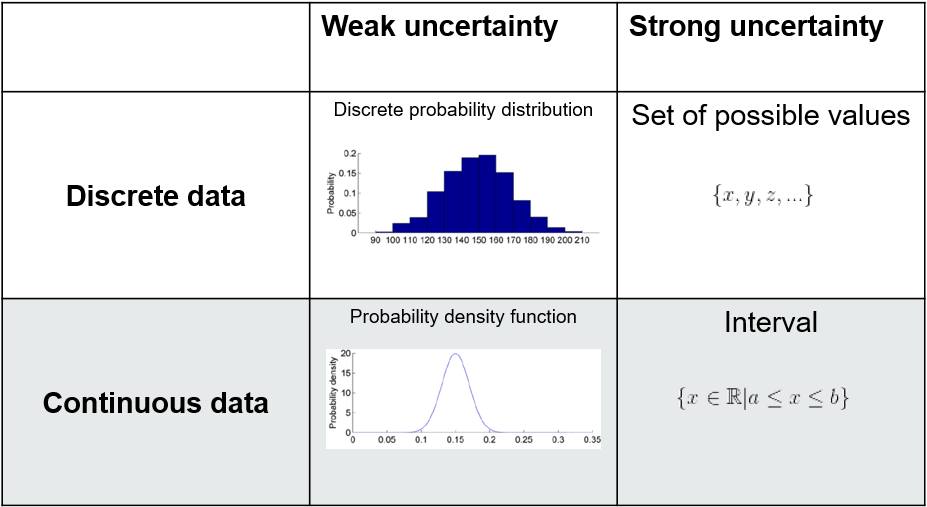}
	\caption{The four different types of uncertainty.}
	\label{fig:unctypes}
\end{figure}

In the case of a discrete attribute, the strong notion of uncertainty consists on a set of possible values assumed by the attribute. In this case, the probability for each possible value is unknown. Vice-versa, in the weak uncertainty scenario we also have a discrete probability distribution defined on that set of values.
In the case of a continuous attribute, the strong notion of uncertainty can be represented with an interval for the variable. Notice that an interval do not indicate a uniform distribution; there is no information on the likelihood of values in it. Vice-versa, in the weak uncertainty scenario we also have a probability density function defined on a certain interval. Figure~\ref{fig:unctypes} summarizes this concepts. This leads to very simple representations of explicit uncertainty.

In this paper we consider only the control flow and time perspective of a process -- namely, the attributes of the events that allow to discover a process model. These are the unique identifier of a process instance (case ID), the timestamp (often represented by the distance from a fixed origin point, e.g. the \emph{Unix Epoch}), and the activity identifier of an event. Case IDs and activities are values chosen from a finite set of possible values; they are discrete variables. Timestamps, instead, are represented by numbers and thus are continuous variables.

We will also describe an additional type of uncertainty, which lays on the event level rather that the attribute level:

\begin{itemize}
	\item \emph{Indeterminate event}: there is a chance that the event did not take place even though it was recorded in the event log. Indeterminate events are indicated with a ? symbol, while determinate (regular) events are marked with a ! symbol.
\end{itemize}

\begin{table}[!htb]
	\centering
	\caption{An example of strongly uncertain trace.}
	\label{table:uncertaintracestrong}
	\begin{tabular}{cccc}
		\textbf{Case ID}                  & \textbf{Timestamp}                                                                                      & \textbf{Activity}                & \multicolumn{1}{l}{\textbf{Indet. event}} \\ \hline
		\multicolumn{1}{|c|}{\{0, 1\}}    & \multicolumn{1}{c|}{2011-12-05T00:00}                                                                   & \multicolumn{1}{c|}{A}           & \multicolumn{1}{c|}{!}                    \\ \hline
		\multicolumn{1}{|c|}{0}           & \multicolumn{1}{c|}{2011-12-07T00:00}                                                                   & \multicolumn{1}{c|}{\{B, C, D\}} & \multicolumn{1}{c|}{!}                    \\ \hline
		\multicolumn{1}{|c|}{0}           & \multicolumn{1}{c|}{\begin{tabular}[c]{@{}c@{}}{[}2011-12-06T00:00,\\ 2011-12-10T00:00{]}\end{tabular}} & \multicolumn{1}{c|}{D}           & \multicolumn{1}{c|}{?}                    \\ \hline
		\multicolumn{1}{|c|}{0}           & \multicolumn{1}{c|}{2011-12-09T00:00}                                                                   & \multicolumn{1}{c|}{\{A, C\}}    & \multicolumn{1}{c|}{!}                    \\ \hline
		\multicolumn{1}{|c|}{\{0, 1, 2\}} & \multicolumn{1}{c|}{2011-12-11T00:00}                                                                & \multicolumn{1}{c|}{E}           & \multicolumn{1}{c|}{?}                    \\ \hline
	\end{tabular}
\end{table}

\begin{table}[!htb]
	\caption{An example of weakly uncertain trace.}
	\label{table:uncertaintraceweak} 
	\centering
	\begin{tabular}{cccc}
		\textbf{Case ID}                       & \textbf{Timestamp}                          & \textbf{Activity}                     & \multicolumn{1}{l}{\textbf{Indet. event}} \\ \hline
		\multicolumn{1}{|c|}{\{0:0.9, 1:0.1\}} & \multicolumn{1}{c|}{2011-12-05T00:00}       & \multicolumn{1}{c|}{A}                & \multicolumn{1}{c|}{!}                    \\ \hline
		\multicolumn{1}{|c|}{0}                & \multicolumn{1}{c|}{2011-12-07T00:00}       & \multicolumn{1}{c|}{\{B:0.7, C:0.3\}} & \multicolumn{1}{c|}{!}                    \\ \hline
		\multicolumn{1}{|c|}{0}                & \multicolumn{1}{c|}{$\mathcal{N}$(2011-12-08T00:00, 2)} & \multicolumn{1}{c|}{D}                & \multicolumn{1}{c|}{?:0.5}                \\ \hline
		\multicolumn{1}{|c|}{0}                & \multicolumn{1}{c|}{2011-12-09T00:00}       & \multicolumn{1}{c|}{\{A:0.2, C:0.8\}} & \multicolumn{1}{c|}{!}                    \\ \hline
		\multicolumn{1}{|c|}{\{0:0.4, 1:0.6\}} & \multicolumn{1}{c|}{2011-12-11T00:00}    & \multicolumn{1}{c|}{E}                & \multicolumn{1}{c|}{?:0.7}                \\ \hline
	\end{tabular}
\end{table}

Examples of strongly and weakly uncertain traces are shown in Tables~\ref{table:uncertaintracestrong} and~\ref{table:uncertaintraceweak} respectively.

\section{Definitions}\label{sec:definitions}
Let us now provide a formal definition of the concept of uncertainty applied to event data.

\begin{definition}[Power Set]
	The power set of a set $A$ is the set of all possible subsets of $A$, and is denoted with $\mathcal{P}(A)$. $\mathcal{P}_{NE}(A)$ denotes the set of all the non-empty subsets of $A$: $\mathcal{P}_{NE}(A) = \mathcal{P}(A)\setminus\{\emptyset\}$.
\end{definition}

\begin{definition}[Multiset]
	A \emph{multiset} is an extension of the concept of set that keeps track of the cardinality of each element. $\bag(A)$ is the set of all multisets over some set $A$. Multisets are denoted with square brackets, e.g. $b = [x, x, y]$.
\end{definition}

\begin{definition}[Sequence]
	Given a set $X$, a finite \emph{sequence} over $X$ of length $n$ is a function $s \in X^* : \{1, \dots, n\} \rightarrow X$, and it is written as $s = \langle s_1, s_2, \dots, s_n\rangle$. Over the sequence $s$ we define $|s| = n$, $s[i] = s_i$ and $x \in s \iff x \in \text{set}(s)$.
\end{definition}

\begin{definition}[Universes]
	Let $\mathcal{U}_E$ be the set of all the \emph{event identifiers}. Let $\mathcal{U}_C$ be the set of all the \emph{case id identifiers}. Let $\mathcal{U}_A$ be the set of all the \emph{activity identifiers}. Let $\mathcal{U}_T$ be the totally ordered set of all the \emph{timestamp identifiers}. Let $\mathcal{U}_O = \{!, ?\}$, where the ``!'' symbol denotes \emph{determined events}, and the ``?'' symbol denotes \emph{indeterminate events}.
\end{definition}

\begin{definition}[Events]
	Let us denote with $\mathcal{E}_C = \mathcal{U}_E \times \mathcal{U}_C \times \mathcal{U}_A \times \mathcal{U}_T$ the universe of \emph{certain events}. $\mathcal{E}_{SU} = \mathcal{U}_E \times \mathcal{P}_{NE}(\mathcal{U}_C) \times \mathcal{P}_{NE}(\mathcal{U}_A) \times \mathcal{P}_{NE}(\mathcal{U}_T) \times \mathcal{U}_O$ is the universe of \emph{strongly uncertain events}. $\mathcal{E}_{WU} =\{(e, f) \in \mathcal{U}_E \times (\mathcal{U}_C \times \mathcal{U}_A \times \mathcal{U}_T \not\to [0,1]) \mid \sum_{(a, c, t) \in dom(f)}f(c, a, t) \leq 1\}$ is the universe of \emph{weakly uncertain events}. Over a strongly uncertain event $(e, c_s, a_s, t_s, u) \in \mathcal{E}_{SU}$ we define the following projection functions: $\pi^{\mathcal{E}_{SU}}_c(e) = c_s$, $\pi^{\mathcal{E}_{SU}}_a(e) = a_s$, $\pi^{\mathcal{E}_{SU}}_t(e) = t_s$ and $\pi^{\mathcal{E}_{SU}}_o(e) = o$.
\end{definition}

\begin{definition}[Event logs]
	A \emph{certain event log} is a set of events $L_C \subseteq \mathcal{E}_C$ such that every event identifier in $L_C$ is unique. A \emph{strongly uncertain event log} is a set of events $L_{SU} \subseteq \mathcal{E}_{SU}$ such that every event identifier in $L_{SU}$ is unique. A \emph{weakly uncertain event log} is a set of events $L_{WU} \subseteq \mathcal{E}_{WU}$ such that every event identifier in $L_{WU}$ is unique. 
\end{definition}

A weakly uncertain event log $L_{WU} \subseteq \mathcal{E}_{WU}$ has a corresponding strongly uncertain event log $\overline{L_{WU}} = L_{SU} \subseteq \mathcal{E}_{SU}$ such that $L_{SU} = \{(e, c_s, a_s, t_s, o) \in \mathcal{E}_{SU} \mid \exists_{(e', f) \in L_{WU}}, e=e' \wedge \\
c_s = \{c \in \mathcal{U}_C \mid \exists_{a, t}((c, a, t) \in dom(f) \wedge f(c, a, t) > 0)\} \wedge \\
a_s = \{a \in \mathcal{U}_A \mid \exists_{c, t}((c, a, t) \in dom(f) \wedge f(c, a, t) > 0)\} \wedge \\
t_s = \{t \in \mathcal{U}_T \mid \exists_{c, a}((c, a, t) \in dom(f) \wedge f(c, a, t) > 0)\} \wedge \\
(o = \:! \iff (\sum_{(c, a, t) \in dom(f)}f(c, a, t)) = 1) \wedge \\ (o = \:? \iff (\sum_{(c, a, t) \in dom(f)}f(c, a, t)) < 1) \}$.

\begin{definition}[Realization of an event log]
	$L_C \subseteq \mathcal{E}_C$ is a \emph{realization} of $L_{SU} \subseteq \mathcal{E}_{SU}$ if and only if:
	\begin{itemize}
		\item For all $(e, c, a, t) \in L_C$ there is a distinct $(e', c_s, a_s, t_s, o) \in L_{SU}$ such that $e'=e$, $a \in a_s$, $c \in c_s$ and $t \in t_s$;
		\item For all $(e, c_s, a_s, t_s, o) \in L_{SU}$ with $o = \:!$ there is a distinct $(e', c, a, t) \in L_C$ such that $e'=e$, $a \in a_s$, $c \in c_s$ and $t \in t_s$.
	\end{itemize}
	$\mathcal{R}_L(L_{SU})$ is the set of all such realizations of the log $L_{SU}$.
\end{definition}
Note that these definition allow us to transform a weakly uncertain log into a strongly uncertain one, and a strongly uncertain one in a set of certain logs.

\section{Conformance Checking on Uncertain Event Data}\label{sec:conformance}
As a preliminary application of process mining over uncertain event data we now focus on conformance checking. Starting from an event log and a process model, conformance checking verifies if the event data in the log conforms to the model, providing a diagnostic of the deviations. Conformance checking serves many purposes, such as checking if process instances follow a specific normative model, assessing if a certain execution log has been generated from a specific model, or verifying the quality of a process discovery technique.

The specific scenario we consider in this paper includes:
\begin{itemize}
	\item Strong uncertainty on the activity;
	\item Strong uncertainty on the timestamp;
	\item Strong uncertainty on indeterminate events.
\end{itemize}
All three can happen concurrently. Table~\ref{table:uncertaintrace} shows such a trace, which we will use as running example. It is worth noticing that the specific case of uncertainty on the case ID causes a problem; since an event can have many possible case IDs, it can belong to different traces. In data format where the event are already aggregated into traces, such as the very common XES standard, this means that the information related to a trace can be \emph{non local} to the trace itself, but can be stored in some other points of the log. We will focus on the problem of uncertainty on the case ID attribute in a future work.

Firstly, we will lay down some simplified notation in order to model in a more compact way the problem at hand.

\begin{definition}[Simple traces and logs]
	$\sigma_C \in \mathcal{U}_A^*$ is a \emph{simple untimed trace}. $\mathcal{T}_C$ denotes the universe of simple untimed traces. $L_C^S \in \bag(\mathcal{T}_C)$ is a \emph{simple untimed log}.
	
	$\sigma_{CT} \in (\mathcal{U}_A \times \mathcal{U}_T)^*$ is a \emph{simple timed trace} if and only if $\sigma_{CT} = \langle (a_1, t_1), (a_2, t_2),\\ \dots, (a_n, t_n) \rangle$ and $\forall_{1 \leq i < j \leq n}, t_i < t_j$. $\mathcal{T}_{CT}$ denotes the universe of simple timed traces. $L_{CT}^S \in \bag(\mathcal{T}_{CT})$ is a \emph{simple timed log}.
	
	$\sigma_U \in \mathcal{P}(\mathcal{U}_E \times \mathcal{P}_{NE}(\mathcal{U}_A) \times \mathcal{U}_T \times \mathcal{U}_T \times \mathcal{U}_O)$ is a \emph{simple uncertain trace} if for all $(e, a_s, t_{min}, t_{max}, u) \in \sigma_U$, $t_{min} < t_{max}$ and all the event identifiers are unique. $\mathcal{T}_U$ denotes the universe of simple uncertain traces.  $L_U^S \in \bag(\mathcal{T}_U)$ is a \emph{simple uncertain log} if all the event identifiers in the log are unique. For $e_U^S = (e, a_s, t_{min}, t_{max}, o) \in \sigma_U$ we define the following projection functions: $\pi^{L_U^S}_a(e_U^S) = a_s$, $\pi^{L_U^S}_{t_{min}}(e_U^S) = t_{min}$, $\pi^{L_U^S}_{t_{max}}(e_U^S) = t_{max}$ and $\pi^{L_U^S}_o(e_U^S) = o$.
\end{definition}

\begin{definition}[Realization of a simple trace]
	$\sigma_{CT} \in \mathcal{T}_{CT}$ is a \emph{timed realization} of $\sigma_U \in \mathcal{T}_U$ if and only if:
	\begin{itemize}
		\item For all $(a, t) \in \sigma_{CT}$ there is a distinct $(a_s, t_{min}, t_{max}, u) \in \sigma_U$ such that $a \in a_s$ and $t_{min} \leq t \leq t_{max}$;
		\item For all $(a_s, t_{min}, t_{max}, u) \in \sigma_U$ with $u = \:!$ there is a distinct $(a, t) \in \sigma_{CT}$ such that $a \in a_s$ and $t_{min} \leq t \leq t_{max}$.
	\end{itemize}
	We denote with $\mathcal{R}_T(\sigma_{U})$ the set of all such timed realizations of the trace $\sigma_{U}$.
	For $\sigma_{CT} \in \mathcal{T}_{CT}$ we denote with $\pi_A(\sigma_{CT})$ the simple untimed trace $\sigma_C \in \mathcal{T}_C$ such that $|\sigma_{CT}|=|\sigma_C|=n$ and for all $1 \leq i \leq n$, $\sigma_{CT}(i) = (a, t)$: $\sigma_C(i) = a$.
	For $\sigma_U \in \mathcal{T}_U$, $\mathcal{R}(\sigma_U) = \{\pi_A(\sigma_{CT}) \mid \sigma_{CT} \in \mathcal{R}_T(\sigma_{U})\}$ is the set of all (untimed) \emph{realizations} of $\sigma_U$.
\end{definition}

These simplified traces and logs can be related to the more general framework described in the previous section through the following transformation: let $L_{SU} \subseteq \mathcal{E}_{SU}$ be a strongly uncertain log and let $g \colon \mathcal{U}_E \not\to \mathcal{U}_C$ be a function mapping events onto cases such that $dom(g) = \{e \mid (e, c_s, a_s, t_s, u) \in L_{SU}\}$ and for all $(e, c_s, a_s, t_s, u) \in L_{SU}$, $g(e) \in c_s$. Thus, for $c \in rng(g)$, $g^{-1}(c) = \{e \in \mathcal{U}_E \mid g(e) = c\}$. The simple uncertain event log defined by $g$ on $L_{SU}$ is $L^S_U = [\{(e, \pi^{\mathcal{E}_{SU}}_a(e), min(\pi^{\mathcal{E}_{SU}}_t(e)), max(\pi^{\mathcal{E}_{SU}}_t(e)), \pi^{\mathcal{E}_{SU}}_o(e)) \mid e \in g^{-1}(c)\} \mid c \in rng(g)]$.

\begin{figure}
	\centering
	\includegraphics[width=1\columnwidth]{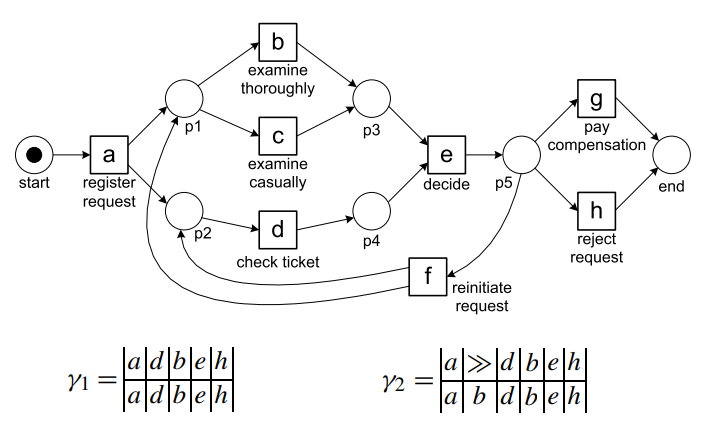}
	\caption{Example of alignments on a model. The alignment $\gamma_1$ shows that the trace $\langle a,d,b,e,h \rangle$ is perfectly fitting the model. The alignment $\gamma_2$ shows that the trace $\langle a,b,d,b,e,h \rangle$ is misaligned with the model in one point.}
	\label{fig:esconformance}
\end{figure}

The conformance checking algorithm that we are applying in this paper is based on \emph{alignments}. Introduced by Adriansyah~\cite{adriansyah2014aligning}, conformance checking through alignments finds deviations between a trace and a Petri net model of a process by creating a correspondence between the sequence of activities executed in the trace and the firing of the transitions in the Petri net. An example of alignments is given in Figure~\ref{fig:esconformance}.

\begin{definition}[System Net]
	A system net is a tuple $SN=(P, T, F, l, M_{init},\\ M_{final})$ with $P$ the set of places, $T$ the set of transitions,	$P \cap T = \emptyset$, $F \subseteq (P \times T) \cup (T \times P)$ the flow relation and $l \in T \not\rightarrow \mathcal{U}_A$ a labeling function over transitions. A marking $M \in \bag(P)$ is a multiset of places; $M_{init} \in \bag(P)$ is the initial marking of the net, and $M_{final} \in \bag(P)$ is the final marking of the net. $\mathcal U_{SN}$ is the \emph{universe of system nets}.
\end{definition}

A system net $SN$ defines a directed graph with nodes $P\cup T$ and edges $F$.
For any $x\in P\cup T$, $\prenet{}{x}=\{y\mid (y,x)\in F\}$ denotes the set of input nodes and
$\postnet{}{x}=\{y\mid (x,y)\in F\}$ denotes the set of output nodes. A transition $t \in T$ is \emph{enabled} in marking $M$ of net $SN$, denoted as $(SN,M)[t\rangle$, if each of its input places $\pre t$ contains at least one token. An enabled transition $t$ may \emph{fire}, i.e., one token is removed from each of the input places $\pre t$ and
one token is produced for each of the output places $\post t$. If $t \notin \mi{dom}(l)$, it is called \emph{invisible}. To indicate invisible transitions we use the placeholder symbol $\tau$; by definition $\tau \notin \mi{dom}(l)$.
An occurrence of visible transition $t \in \mi{dom}(l)$ corresponds to observable activity $l(t)$. Given a system net, $\phi(\mi{SN})$ is the set of all possible \emph{visible} activity sequences, i.e. the labels of complete firing sequences starting in $M_{\mi{init}}$ and ending in $M_{\mi{final}}$ projected onto the set of observable activities. Given the set of activity sequences $\phi(\mi{SN})$ obtainable via complete firing sequences on a certain system net, we can define a perfectly fitting event log as a set of traces which activity projection is contained in $\phi(\mi{SN})$.

These definitions allow us to build \emph{alignments} in order to compute the fitness of trace on a certain model. An alignment is a correspondence between a sequence of activities (extracted from the trace) and a sequence of transitions with the relative labels (fired in the model while replaying the trace). The first sequence indicates the ``moves in the log'' and the second indicates the ``moves in the model''. If a move in the model cannot be mimicked by a move in the log, then a ``$\nomove$'' (``no move'') appears in the top row; conversely, if a move in the log cannot be mimicked by a move in the model, then a ``$\nomove$'' (``no move'') appears in the bottom row.``no moves'' not corresponding to invisible transitions point to
deviations between model and log. A \emph{move} is a pair $(x,(y,t))$ where the first element refers to the log and the second element to the model. A ``$\nomove$'' in the first element of the pair indicates a move on model, in the second element it indicates a move on log.

An alignment is a sequence of moves such that after removing all ``$\nomove$'' symbols,
the top row corresponds to a trace in the log and the bottom row corresponds
to a firing sequence starting in $M_{\mi{init}}$ and ending $M_{\mi{final}}$. Notice that if $t \notin \mi{dom}(l)$ is an invisible transition, the activation of $t$ is indicated by a ``$\nomove$'' on the log in correspondence of $t$ and the placeholder label $\tau$.
Hence, the middle row corresponds to a visible path when ignoring the $\tau$ steps. Figure~\ref{fig:esconformance} shows a model with two examples of alignments, one of a fitting trace and the other of a non-fitting trace.

\begin{definition}[Alignment]
	Let $\sigma_C \in L_C$ be a trace and $t \in \phi_f(\mi{SN})$ a complete firing sequence of system net $\mi{SN}$.
	An \emph{alignment} of $\sigma_C$ and $t_*$ is a sequence of moves $\gamma \in {A_{M}}^*$
	such that the projection on the first element (ignoring $\nomove$) yields $\sigma$
	and the projection on the last element (ignoring $\nomove$ and transition labels) yields $t$.
\end{definition}

A trace and a model can have several possible alignments. In order to select the most appropriate one, we introduce a function that associate a \emph{cost} to undesired moves - the ones associated with deviations.

\begin{definition}[Cost of Alignment]\label{def:alcosts}
	Cost function $\delta \in {A_{LM}} \rightarrow \Nat$ assigns costs to legal moves.
	The \emph{cost} of an alignment $\gamma \in {A_{LM}}^*$ is the sum of all costs:
	$\delta(\gamma) = \sum_{(x,y)\in \gamma} \delta(x,y)$.
\end{definition}

In this paper we use a standard cost function $\delta_S$
that assigns cost zero to synchronous moves and moves on invisible transitions, and unit costs to moves on log or moves on model.

\begin{definition}[Optimal Alignment]
	Let $L_C \in \bag(\mathcal{T}_C)$ be a simple untimed event log and let $\mi{SN}\in {\cal U}_{\mi{SN}}$ be a system net with $\phi(\mi{SN}) \neq \emptyset$.
	\begin{itemize}
		\item For $\sigma_C \in L_C$, we define:
		$\Gamma_{\sigma_C,SN} = \{ \gamma \in  {A_{LM}}^* \mid \exists_{t_* \in \phi_f(\mi{SN})} \  \gamma \ \mathit{is} \ \mathit{an} \ \allowbreak \mathit{alignment} \ \mathit{of} \ \sigma_C \ \mathit{and} \ t_* \}$.
		
		\item An alignment $\gamma \in \Gamma_{\sigma_C,SN}$ is \emph{optimal} for trace $\sigma_C \in L_C$ and system net $\mi{SN}$
		if for any $\gamma' \in  \Gamma_{\sigma_C,M}$: $\delta(\gamma') \geq \delta(\gamma)$.
		
		\item $\lambda_{SN} \in \mathcal{E}^* \rightarrow {A_{LM}}^* $ is
		a deterministic mapping that assigns any trace $\sigma_C$ to an optimal alignment, i.e., $\lambda_{SN}(\sigma_C) \in \Gamma_{\sigma_C,SN}$ and $\lambda_{SN}(\sigma_C)$ is optimal.
		
		\item $\mi{costs}(L_C,\mi{SN},\delta) = \sum_{\sigma_C \in L} \delta(\lambda_{SN}(\sigma_C))$ are the \emph{misalignment costs} of the whole event log.
	\end{itemize}
\end{definition}

Depending on the possible values for $a_s$, $t_{min}$, $t_{max}$, and $u$ there are multiple possible realizations of a trace. This means that, given a model, a simple uncertain trace could be fitting for certain realizations, but non-fitting for others. The question we are interested in answering is: given a simple uncertain trace and a Petri net process model, is it possible to find an \emph{upper and lower bound} for the conformance score? More formally, when usually we are interested in the optimal alignments (the ones with the minimal cost), we are now interested in the minimum and maximum cost of alignments in the realization set of a simple uncertain trace.

\begin{table}[]
	\caption{The uncertain trace used as running example for the application of Conformance Checking on uncertainty.}
	\label{table:uncertaintrace}
	\centering
	\begin{tabular}{cccc}
		\textbf{Case ID}        & \textbf{Timestamp}                                                                                                     & \textbf{Activity}             & \multicolumn{1}{l}{\textbf{Indet. event}} \\ \hline
		\multicolumn{1}{|c|}{0} & \multicolumn{1}{c|}{2011-12-05T00:00}                                                                         & \multicolumn{1}{c|}{A}        & \multicolumn{1}{c|}{!}                    \\ \hline
		\multicolumn{1}{|c|}{0} & \multicolumn{1}{c|}{2011-12-07T00:00}                                                                         & \multicolumn{1}{c|}{\{B, C\}} & \multicolumn{1}{c|}{!}                    \\ \hline
		\multicolumn{1}{|c|}{0} & \multicolumn{1}{c|}{\begin{tabular}[c]{@{}c@{}}[2011-12-06T00:00\\ 2011-12-10T00:00]\end{tabular}} & \multicolumn{1}{c|}{D}        & \multicolumn{1}{c|}{!}                    \\ \hline
		\multicolumn{1}{|c|}{0} & \multicolumn{1}{c|}{2011-12-09T00:00}                                                                         & \multicolumn{1}{c|}{\{A, C\}} & \multicolumn{1}{c|}{!}                    \\ \hline
		\multicolumn{1}{|c|}{0} & \multicolumn{1}{c|}{2011-12-11T00:00}                                                                         & \multicolumn{1}{c|}{E}        & \multicolumn{1}{c|}{?}                    \\ \hline
	\end{tabular}
\end{table}

\begin{definition}[Upper and Lower Bound on Alignment Cost for a Trace]
	Let $\sigma_U \in \mathcal{T}_U$ be a simple uncertain trace, and let $SN \in \mathcal{U}_{SN}$ be a system net. The \emph{upper bound for the alignment cost} is a function $\delta_{max} \colon \mathcal{T}_U \to \mathbb{N}$ such that $\delta_{max}(\sigma_U) = \max_{\sigma_C \in \mathcal{R}(\sigma_U)} \lambda_{SN}(\sigma_C)$. The \emph{lower bound for the alignment cost} is a function $\delta_{min} \colon \mathcal{T}_U \to \mathbb{N}$ such that $\delta_{min}(\sigma_U) = \min_{\sigma_C \in \mathcal{R}(\sigma_U)} \lambda_{SN}(\sigma_C)$.
\end{definition}

A simple way to compute the upper and lower bounds for the cost of an uncertain trace is using a bruteforce approach: enumerating the possible realizations of the trace, then searching for the costs of optimal alignments for all the realizations, and picking the minimum and maximum as bounds.

The technique to compute the optimal alignment~\cite{adriansyah2014aligning} is as follows. Firstly, it creates an \emph{event net}, a sequence-structured system net able to replay only the trace to align. The transitions in the event net have labels corresponding to the activities in the trace. Then, a \emph{product net} should be computed; it is the union of the event net and the model together with synchronous transitions added. These additional transitions are paired with transitions in the event net and in the process model that have the same label; they are then connected with arcs from the input places and to the output places of those transitions. The product net is able to represent moves on log, moves on model and synchronous moves by means of firing transitions: the transitions of the event net correspond to moves on log, the transitions of the process model correspond to moves on model, the added synchronous transitions correspond to synchronous moves. The union of the initial and final markings of the event net and the process model constitute respectively the initial and final marking of the product net: every complete firing sequence on the product net corresponds to a possible alignment. Lastly, the product net is translated to a state space, and a state space exploration via the $\mathbb{A}^*$ algorithm is performed in order to find the complete firing sequence that yields the lowest cost.

Let us define formally the construction of the event net and the product net:

\begin{definition}[Event Net]
	Let $\sigma_C \in \mathcal{T}_C$ be a simple untimed trace. The \emph{event net} $en: \mathcal{T}_C \to \mathcal{U}_{SN}$ of $\sigma_C$ is a system net $en(\sigma_C) = (P, T, F, l, M_{init}, M_{final})$ such that:
	\begin{itemize}
		\item $P = \{p_i \mid 1 \leq i \leq |\sigma_C|+1 \}$,
		\item $T = \{t_i \mid 1 \leq i \leq |\sigma_C| \}$,
		\item $F = \bigcup_{1 \leq i \leq }\{(p_i, t_i), (t_i, p_{i + 1})\}$
		\item $l \colon T \to \mathcal{U}_A$ such that for all $1 \leq i \leq |\sigma_C|$, $l(t_i) = \sigma_C[i]$,
		\item $M_{init} = \{ p_1 \}$,
		\item $M_{final} = \{ p_{|P|} \}$.
	\end{itemize}
\end{definition}

\begin{definition}[Product of two Petri Nets]
	Let $S_1 = (P_1, T_1, F_1, l_1, M_{init_1},\\ M_{final_1})$ and $S_2 = (P_2, T_2, F_2, l_2, M_{init_2}, M_{final_2})$ be two system nets. The \emph{pro\-duct net} of $S_1$ and $S_2$ is the system net $S = S_1 \otimes S_2 = (P, T, F, l, M_{init}, M_{final})$ such that:
	\begin{itemize}
		\item $P = P_1 \cup P_2$,
		\item $T \subseteq (T_1 \cup \{\nomove\} \times T_2 \cup \{\nomove\})$ such that $T = \{(t_1, \nomove) \mid t_1 \in T_1 \} \cup \{(\nomove, t_2) \mid t_2 \in T_2 \} \cup \{(t_1, t_2) \in (T_1 \times T_2) \mid l_1(t_1) = l_2(t_2) \neq \tau \}$,
		\item $F \subseteq (P \times T) \cup (T \times P)$ such that \\
		$F = \{(p, (t, \nomove)) \mid p \in P_1 \wedge t \in T_1 \wedge (p, t) \in F_1 \} \cup
		\\
		\{((t, \nomove), p) \mid t \in T_1 \wedge p \in P_1 \wedge (t, p) \in F_1 \} \cup \\ 
		\{(p, (t, \nomove)) \mid p \in P_2 \wedge t \in T_2 \wedge (p, t) \in F_2 \} \cup \\
		\{((t, \nomove), p) \mid t \in T_2 \wedge p \in P_2 \wedge (t, p) \in F_2 \} \cup \\
		\{(p, (t_1, t_2)) \mid p \in P_1 \cup P_2 \wedge (t_1, t_2) \in T \cap (T_1 \times T_2) \} \cup \\
		\{((t_1, t_2), p) \mid p \in P_1 \cup P_2 \wedge (t_1, t_2) \in T \cap (T_1 \times T_2) \}$
		\item $l \colon T \to \mathcal{U}_A$ such that for all $(t_1, t_2) \in T$, $l((t_1, t_2)) = l_1(t_1)$ if $t_2 = \nomove$, $l((t1, t2)) = l_2(t_2)$ if $t_1 = \nomove$, and $l((t_1, t_2)) = l_1(t_1)$ otherwise,
		\item $M_{init} = M_{init_1} \uplus M_{init_2}$,
		\item $M_{final} = M_{final_1} \uplus M_{final_2}$.
	\end{itemize}
\end{definition}

We now present a technique which improves the performance of calculating the lower bound for conformance cost over using a bruteforce method. We will produce a version of the event net that embeds the possible behaviors of the uncertain trace. We define a \emph{behavior net}, a Petri net that can replay all and only the realizations of an uncertain trace.

In order to obtain such a Petri net we first built a directed graph representing the uncertain trace as an intermediate step. We also need to present the concept of \emph{transitive reduction}: given a directed graph $G$, its transitive reduction $G'$ is a graph with the same set of vertices, the same reachability between vertices, and a minimal number of arcs, such that every pair of vertices is connected by at most one path. The transitive reduction of a directed acyclic graph always exists and is unique~\cite{aho1972transitive}.

We can then define the \emph{behavior graph}, which contains a vertex for each uncertain event in the trace and contains an edge between two vertices if the corresponding uncertain events may happen one directly after the other.

\begin{definition}[Behavior Graph]
	Let $\sigma_U \in \mathcal{T}_U$ be a simple uncertain trace. A \emph{behavior graph} $bg \colon \mathcal{T}_U \to \mathcal{U}_G$ is the transitive reduction of a  directed graph $(V, E)$, where $V$ is the set of vertices and $E \subseteq V \times V$ is the set of directed edges, such that:
	\begin{itemize}
		\item $V = \sigma_U$
		\item $E = \{(v_1, v_2) \mid v_1 \in V, v_2 \in V, \pi^{L_U^S}_{t_{max}}(v_1) < \pi^{L_U^S}_{t_{min}}(v_2)\}$
	\end{itemize}
\end{definition}

\begin{figure}
	\centering
	\begin{minipage}{0.45\textwidth}
		\centering
		\includegraphics[width=0.9\textwidth, height=0.35\textheight, keepaspectratio]{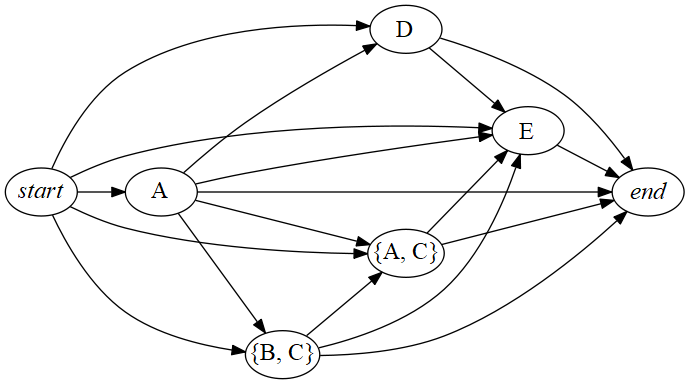} 
		\caption{The behavior graph of the trace in Table~\ref{table:uncertaintrace} before applying the transitive reduction.}
		\label{fig:graphcomp}
	\end{minipage}\hfill
	\begin{minipage}{0.45\textwidth}
		\centering
		\includegraphics[width=0.9\textwidth, height=0.35\textheight, keepaspectratio]{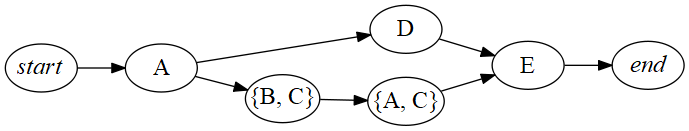} 
		\caption{The same behavior graph after the transitive reduction.}
		\label{fig:graphred}
	\end{minipage}
\end{figure}

The behavior graph provides a structured representation of the uncertainty on the timestamp: when a specific vertex has two or more outbound edges, the events corresponding to the destination vertices can occur in any order, concurrently with each other. The property of $\mathcal{U}_T$ of being totally ordered and the property $t_{min} < t_{max}$ of all simple uncertain traces ensure that the behavior graph is acyclic both before and after the transitive reduction. We can see the result on the example trace in Figures~\ref{fig:graphcomp} and~\ref{fig:graphred}.

We then obtain a \emph{behavior net} by replacing every vertex in the behavior graph with one or more transitions in a XOR configuration, each representing an activity contained in the $\pi_\mathcal{A}$ set of the corresponding uncertain event. The edges of the behavior graph become connection through places in the behavior net.

\begin{definition}[Behavior Net]
	Let $\sigma_U \in \mathcal{T}_U$ be a simple uncertain trace, and let $\text{bg}(\sigma_U) = (V, E)$ be the corresponding behavior graph. A \emph{behavior net} $bn \colon \mathcal{T}_U \to \mathcal{U}_{SN}$ is a system net $bn(\sigma_U) = (P, T, F, l, M_{init}, M_{final})$ such that:
	\begin{itemize}
		\item $T = \{(v, a) \mid v \in V \wedge a \in \pi^{L_U^S}_a(v)\} \cup \{(v, \tau) \mid v \in V \wedge a \in \pi^{L_U^S}_a(v) \wedge \pi^{L_U^S}_o(v) = \:?\}$
		\item $P = E \cup \{\text{start}, \text{end}\}$
		\item $F = \{((v_1, a),(v_1, v_2)) \mid (v_1, a) \in T, (v_1, v_2) \in E\} \cup \\
		\{((v_1, v_2),(v_2, a)) \mid (v_1, v_2) \in P, (v_2, a) \in E\} \cup \\
		\{(\text{start}, (v, a)) \mid (v, a) \in T \wedge \forall_{v_* \in V} \pi^{L_U^S}_{t_{min}}(v) < \pi^{L_U^S}_{t_{min}}(v_*) \} \cup \\
		\{((v, a), \text{end}) \mid (v, a) \in T \wedge \forall_{v_* \in V} \pi^{L_U^S}_{t_{max}}(v) > \pi^{L_U^S}_{t_{max}}(v_*) \}$
		\item $l = \{((v, a), a) \mid (v, a) \in T \wedge a \neq \tau\}$
		\item $M_{\text{init}} = \{\text{start}\}$
		\item $M_{\text{final}} = \{\text{end}\}$
	\end{itemize}
\end{definition}

\begin{figure}
	\centering
	\includegraphics[width=1\columnwidth]{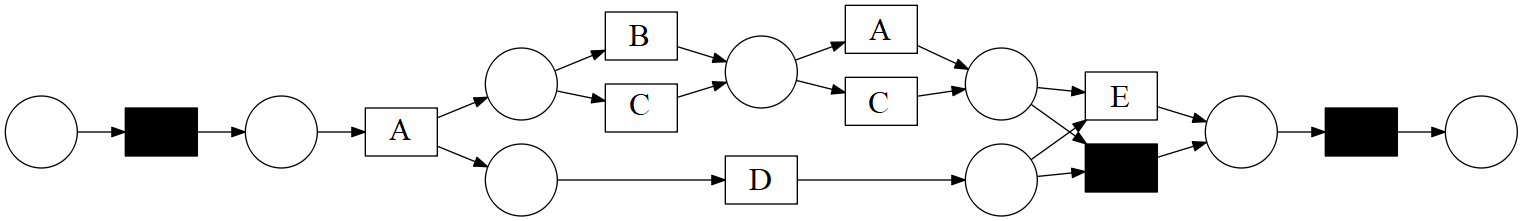}
	\caption{The behavior net corresponding to the uncertain trace in Table~\ref{table:uncertaintrace}.}
	\label{fig:behnet}
\end{figure}

In Figure~\ref{fig:behnet} we can see the behavior net corresponding to the uncertain trace in Table 2. It is important to notice that every set of edges in the behavior graph with the same source vertex generate an AND split in the behavior net, and a set of edges with the same destination vertex generate an AND join. At the same time, the transitions which labels correspond to different possible activities in an uncertain event will appear in a XOR construct inside the behavior net.

This means that every set of events which timestamps allow for overlapping will be represented in the behavior net by transitions inside an AND construct, and will then allow to execute in the net all the possible sequences of events obtained choosing a possible value for the uncertain timestamp attribute. In the same fashion, an event with uncertainty on the activity will be represented by a number of transitions in a XOR construct, that allows to replay any possible choice for the activity attribute. It follows that, by construction, for a certain simple uncertain trace $\sigma_U$ we have that $\phi(bn(\sigma_U)) = \mathcal{R}(\sigma_U)$.

We can use the behavior net of an uncertain trace $\sigma_U$ in lieu of the event net to compute alignments with a model $SN \in \mathcal{U}_{SN}$; the search algorithm returns an optimal alignment, a sequence of moves $(x, (y, t))$ with $x \in \mathcal{U}_A$, $y \in \mathcal{U}_A$ and $t$ transition of the model $SN$. After removing all ``$\nomove$'' symbols, the sequence of first elements of the moves will describe a complete firing sequence $\sigma_X$ of the behavior net. Since $\sigma_X$ is complete, $\sigma_X \in \phi(bn(\sigma_U))$ and, thus, $\sigma_X \in \mathcal{R}(\sigma_U)$. It follows that $\sigma_X$ is a realization of $\sigma_U$, and the search algorithm ensures that $\sigma_X$ is a realization with optimal conformance cost for the model $SN$: $\delta(\lambda_{SN}(\sigma_X)) = \min_{\sigma_C \in \mathcal{R}(\sigma_U)} \lambda_{SN}(\sigma_C) = \delta_{min}(\sigma_U)$.

\section{Experiments}\label{sec:experiments}
The technique to compute conformance for strongly uncertain traces and to create the behavior net hereby described has been implemented for testing, using the code for alignments already provided in the process mining Python library PM4Py~\cite{pm4py}. Uncertainty has been represented in the XES standard through meta-attributes and constructs such as lists, such that any XES importer can read an uncertain log file. The algorithm was designed to be fully compatible with non-uncertain XES event logs; the meta-attributes for uncertainty were designed to be partly compatible with other process mining algorithms -- meta-attributes describing the possible values for an uncertain activity or the interval of an uncertain timestamp can also specify a ``fallback value'' that other process mining software will read as (certain) activity or timestamp value.

Two experiments were run: the first to inspect the bounds for conformance score as increasingly more uncertainty is added to an event log; the other test assesses the difference in performance between the bruteforce method and the behavior net. We ran the tests on synthetic event logs where we added uncertainty. This way we can control the amounts of uncertainty in event data. Through the ProM plugin \emph{``Generate block-structured stochastic Petri nets''} we generated Petri nets of different sizes in terms of number of transitions $n$; then, we used PM4Py in order to generate event logs adding uncertainty to attributes. Activities and timestamps are uncertain with probability $p$; also, events have a chance to be indeterminate with probability $p$.

The pipeline for the first experiment was the following: we generated a model with $n$ = 10; we executed the model to obtain 250 traces; we added deviations to events (every event has 20\% chances to have the wrong activity, every pair of consecutive events have 20\% chances of having the timestamps swapped, every trace has 40\% chances to have an additional event). We then added uncertainty to the events: each event has probability $p$ of having two possible values for the activity, probability $p$ of having an uncertain timestamp, and probability $p$ to be an indeterminate event. We calculated the bounds on conformance cost of the log so generated and repeated the procedure for increasing values of $p$. 

\begin{figure}
	\centering
	\includegraphics[width=1\columnwidth]{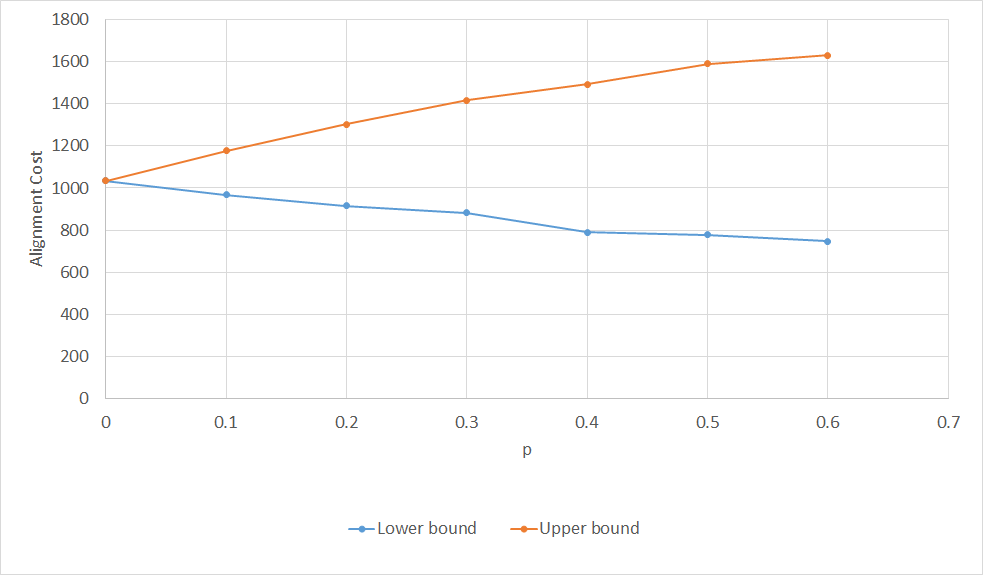}
	\caption{The change in lower and upper bound for conformance checking of an event log with increasing probability of having uncertainty on event data.}
	\label{fig:deviation}
\end{figure}

Figure~\ref{fig:deviation} shows the results. We can see that the cost shows the expected behavior: at $p$ = 0 the two bounds coincide, since the traces are certain and have only one realization. Conversely, the log with $p$ = 0.6 has a total of 1629 deviations on the worst case scenario (6.52 on average per trace), and 747 deviations in the best case scenario (2.99 on average per trace); a process that includes traces with comparable uncertainty has thus a huge difference in behavior between the best case scenario and worst case scenario. The evaluation of the best and worst case scenario for uncertain traces can give useful indications to a business user on the parts of the process where there is a high risk of deviation, in order to enhance them.

The second experiment concerns the performance of calculating the lower bound of the cost via the behavior net versus the bruteforce method of separately listing all the realizations of an uncertain trace, evaluating all of them through alignments, then picking the best value. We used a constant value of $p = 0.2$ and logs of 100 traces for this test, with progressively increasing values of $n$.

\begin{figure}
	\centering
	\includegraphics[width=1\columnwidth]{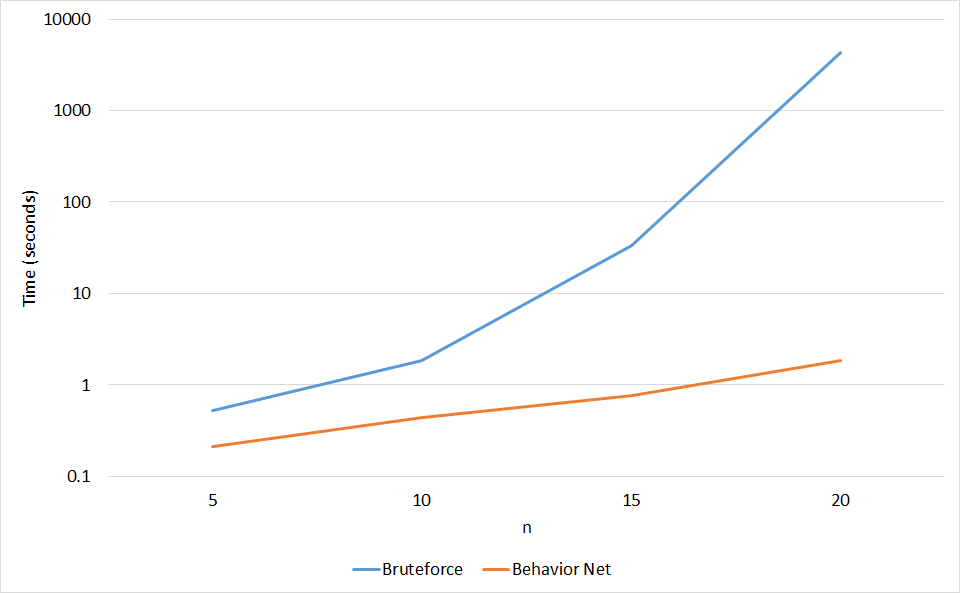}
	\caption{Effect on time performance of calculating the lower bound for conformance cost with the bruteforce method vs. the behavior net.}
	\label{fig:performance}
\end{figure}

Figure~\ref{fig:performance} summarizes the results. As the diagram shows, the difference in time between the two methods tends to diverge quickly even on a logarithmic scale. With $n$ = 5, the behavior net provides the lower bound in 40.4\% of the time required by the bruteforce method. For $n$ = 20, the largest model we could test, the behavior net takes 0.04\% of the time needed by the bruteforce method. This shows a very large improvement in the computing time for the lower bound computation, so the best case scenario for the conformance cost of an uncertain trace can be obtained efficiently thanks to the structural properties of the behavior net.

\section{Conclusion}\label{sec:conclusion}
As the need of quickly and effectively analyze process data has arisen in the recent past and is growing to this day, many new types of information regarding events are recorded; this calls for new techniques able to provide an adequate interpretation of the new data. In this paper we presented a new paradigm for process mining applied to event data: explicit uncertainty. We described the possible form it can assume, building a taxonomy of different types of uncertainty. We then designed a formal mathematical infrastructure to define the various flavors of uncertainty shown in the taxonomy. Then, in order to assess the practical applications of the uncertainty framework, we applied it to a well consolidated technique for conformance checking: aligning data to a reference Petri net. The results can provide insights on the possible violations of process instances recorded with uncertainty against a normative model. The behavior net provides an efficient way to compute the lower bound for the conformance cost -- i.e. the best case scenario for conformity of uncertain process data -- with a large improvement on time performance with respect to a bruteforce procedure.

The approaches shown here can be extended in a number of ways. An important step in this line of research is assessing the technique on real-life logs. From a performance perspective, to improve the usability of alignments over uncertainty we shall optimize the computation of the upper bound of the conformance cost. Another natural continuation of this work is extending the conformance checking technique to logs with weak uncertainty. Many possibilities can be pursued to broaden the concept of uncertainty on different process mining methods: for example, discovering a Petri net from an uncertain event log, or develop techniques to mine Petri nets that embed uncertainty information about the process.

\bibliographystyle{splncs04}
\bibliography{mining-uncertain-event-data-in-process-mining}

\begin{thebibliography}{10}
\providecommand{\url}[1]{\texttt{#1}}
\providecommand{\urlprefix}{URL }
\providecommand{\doi}[1]{https://doi.org/#1}

\bibitem{pm4py}
Pm4py, \url{http://pm4py.pads.rwth-aachen.de/}

\bibitem{van2011process}
Van~der Aalst, W., Adriansyah, A., De~Medeiros, A.K.A., Arcieri, F., Baier, T.,
  Blickle, T., Bose, J.C., Van Den~Brand, P., Brandtjen, R., Buijs, J., et~al.:
  Process mining manifesto. In: International Conference on Business Process
  Management. pp. 169--194. Springer (2011)

\bibitem{adriansyah2014aligning}
Adriansyah, A.: Aligning observed and modeled behavior  (2014)

\bibitem{aho1972transitive}
Aho, A.V., Garey, M.R., Ullman, J.D.: The transitive reduction of a directed
  graph. SIAM Journal on Computing  \textbf{1}(2),  131--137 (1972)

\bibitem{carmona2018conformance}
Carmona, J., van Dongen, B., Solti, A., Weidlich, M.: Conformance Checking:
  Relating Processes and Models. Springer (2018)

\bibitem{conforti2018timestamp}
Conforti, R., La~Rosa, M., ter Hofstede, A.: Timestamp repair for business
  process event logs (2018), \url{http://hdl.handle.net/11343/209011},
  [preprint]

\bibitem{conforti2017filtering}
Conforti, R., La~Rosa, M., ter Hofstede, A.H.: Filtering out infrequent
  behavior from business process event logs. IEEE Transactions on Knowledge and
  Data Engineering  \textbf{29}(2),  300--314 (2017)

\bibitem{han2011data}
Han, J., Pei, J., Kamber, M.: Data mining: concepts and techniques. Elsevier
  (2011)

\bibitem{sani2017improving}
Sani, M.F., van Zelst, S.J., van~der Aalst, W.M.: Improving process discovery
  results by filtering outliers using conditional behavioural probabilities.
  In: International Conference on Business Process Management. pp. 216--229.
  Springer (2017)

\bibitem{sani2018repairing}
Sani, M.F., van Zelst, S.J., van~der Aalst, W.M.: Repairing outlier behaviour
  in event logs. In: International Conference on Business Information Systems.
  pp. 115--131. Springer (2018)

\bibitem{suriadi2017event}
Suriadi, S., Andrews, R., ter Hofstede, A.H., Wynn, M.T.: Event log
  imperfection patterns for process mining: Towards a systematic approach to
  cleaning event logs. Information Systems  \textbf{64},  132--150 (2017)

\bibitem{wang2015cleaning}
Wang, J., Song, S., Lin, X., Zhu, X., Pei, J.: Cleaning structured event logs:
  A graph repair approach. In: Data Engineering (ICDE), 2015 IEEE 31st
  International Conference on. pp. 30--41. IEEE (2015)

\bibitem{van2018filtering}
van Zelst, S.J., Sani, M.F., Ostovar, A., Conforti, R., La~Rosa, M.: Filtering
  spurious events from event streams of business processes. In: International
  Conference on Advanced Information Systems Engineering. pp. 35--52. Springer
  (2018)

\end{thebibliography}

\end{document}